# Minimal Feature Analysis for Isolated Digit Recognition for varying encoding rates in noisy environments

Muskan Garg, Panjab University, India
Naveen Aggarwal, Panjab University, India
muskanphd@gmail.com , navagg@gmail.com

**ABSTRACT**

This research work is about recent development made in speech recognition. In this research work, analysis of isolated digit recognition in the presence of different bit rates and at different noise levels has been performed. This research work has been carried using audacity and HTK toolkit. Hidden Markov Model (HMM) is the recognition model which was used to perform this experiment. The feature extraction techniques used are Mel Frequency Cepstrum coefficient (MFCC), Linear Predictive Coding (LPC), perceptual linear predictive (PLP), mel spectrum (MELSPEC), filter bank (FBANK). There were three types of different noise levels which have been considered for testing of data. These include random noise, fan noise and random noise in real time environment. This was done to analyse the best environment which can used for real time applications. Further, five different types of commonly used bit rates at different sampling rates were considered to find out the most optimum bit rate.

# 1. INTRODUCTION

Fundamental speech units are called **phonemes**. Two adjacent and three adjacent phonemes are called diphones and triphones respectively. Larger speech units are framed by using set of phonemes called words. During pronunciation, the effect of two phonemes on each other when spoken continuously is called **co-articulation** affect. Words are modeled by two components namely acoustic model and language model. **Acoustic modeling** is the process of constructing statistical representation and its structures automatically from feature vector sequences computed from speech waveform. **Language modeling** is the process of computing the likelihood for possible set of words trained from corpus. Speech corpus is the database of audio files and text transcription. Speech is also used in cell phones (Smartphone) these days. These devices have predefined commands to work on customer speech commands. IVR is used for interaction of humans and computers using voice and DTMF tones via input keypad. Voice over Internet Protocol (VoIP) [4] is audio traffic over internet which is encrypted before transmission by ASR. Speech chain refers to the entire chain being carried from sharing the views/ thoughts of the speaker to the understanding of listener. It has two phase namely speech production and speech perception.

## 1.1 DIGITIZATION

Bit rate is the number of bits per second. Here, audio files are being considered. As per different audio formats, it has been observed that quality of sounds is dependent upon bit rates. The normal audio CD is 16-bit 44.1 kHz Stereo PCM audio. As there is good quality in sound tracks with higher bit rate, higher the bit rate less will be compression. Better quality and higher bit rate are interrelated in context of more bits in signal per second and more data can be examined from the speech signal respectively. Audio formats are usually 'lossy', as the data can be lost after being recorded. Storing the high level quality speech sounds accompanies capturing of more data. Thus, storage is at higher bit rate, more will be data, lesser will be compression i.e. more retention and less loss of data. Table 1 shows different sampling rates.

**Table 1: Standard formats of Audio**

| Type | Bit Resolution | Sampling Rate | Bit Rate |
|---|---|---|---|
| Telephone (Landline) | 8 – bit | 8000 Hz | 64000 bps |
| VoIP and Improved Telephony | 8 – bit | 16000 Hz | 128 kbps |
| Broadcast Rate (Better than FM Radio) | 16 – bit | 32000 Hz | 512 kbps |
| Compact Disc | 16 – bit | 44100 Hz | 705 kbps |
| High Resolution Audio (DVD) | 24 – bit | 48000 Hz | 1.1 Mbps |

The standard sampling audio rate for DVDs is 96 kHz, for DAT recorders is 48 kHz, for CD is 44.1 kHz for MAC and other desktop computers is 22 kHz to 11 kHz. The bit rate varies as per objective. 8 bit is used for simple less precise signals, 16-bits is used for more precise signals and so on. Bit rate quantizes the sampling rate and introduces precision. More the bit rate better will be precision of data.

During speech recognition PCM signal is used to convert the analog signal to digital signal. PCM signal signifies Pulse Code Modulation, a technique or methods which standardizes the audio signals in Digital Video Disc, Compact Disc, DAT etc. Nyquist frequency is the half of the sampling rate of audio signal. Nyquist theorem states that the sampling rate of the audio signal should be more than the double of the maximum frequency of the audio signal.

Speech signals are trained and stored in speech corpus. Target signals are matched with trained signals. The output obtained is the result of matching. The performance is usually measured in word error rate (WER). According to the performance measurement of the result obtained, the accuracy of the signal is tested.

$$WER = (S+D+I)/N$$

Where S = number of substitutions, D = number of deletions, I = number of insertions, N= number of words of reference.

There are various applications and vast scope to extend the research in the same. There areas include medical research, speaker recognition in forensic laboratories, extracting the information of the subject from the voice, formulating new algorithms or techniques to identify the continuous speaker independent speech [3]. It is used widely in gaming especially in human-computer interaction business games, teaching methodologies, virtual reality, and telephone directory automated enquiry of the customer. Other applications include digital forensics, robotics, language translations, getting the audio lecture in text etc..

Following steps are carried to take speech recognition:

- Recording related voices at different bit rates
- Extracting various audio features at different temporal derivatives
- Preparing various Speech Corpus
- Implementation of different recognition techniques
- Extracting features of target audio signal
- Matching the later features with former ones.
- Obtaining results in the form of word error rate.
- Calculating accuracy with acceptance or rejection of results

## 2. RELATED WORK

However, in 1974 L. R. Rabiner et al. [15] proposed a new algorithm for identifying the endpoints of the spoken words in the presence of environment. Zero crossings and energy are two important factors based on which the author proposed the new algorithm for the same. It has been tested on large number of speakers and thus, is very efficient. On the other hand, in 1981, Lori F. Lamel et al. [16] proposed the hybrid method of detecting the endpoints of the word and found that it bears less than 0.5% of error rate. As there is no separate stage for endpoint detection, the combination of different algorithms has been used in order to examine the endpoint of word in the presence of background sounds being used. In 1984, Marcia A. Bush et a. [19] proposed two techniques in order to segment the syllables in the speech signal. One of them is to automatically segment the audio wave into equal length segments and the other one is to manually segment the acoustic phonetic units. However, both methods gain 96% to 97% of accuracy. Author used vector quantization technique for the same.

The sound being produced by human being is filtered by vocal tracts (teeth, tongue) etc.. The accuracy achieved in determination of shape gives the representation of phonemes. The MFCC represents envelop of short time power spectrum being framed by vocal tract of human being during occurrences. MFCC has been introduced by Davis and Mermelstein in 1980's. Prior to MFCC, LPC was being used. In 2010, Vibha Tiwari [33] performed speech recognition using MFCC and Vector Quantization technique. She has proposed large number of applications where MFCC feature vector extraction technique is used. Also, she has proposed many modifications in the algorithms during her case study on MFCC method. MFCC has been described by her in detail in her writings.

It has been observed that recognition of isolated words, its boundaries and continuous words are being carried on single bit rate after multiple research work are being referred. So, this research work have been carried on various bit rates and sampling rates used to get speech samples. Speech samples are further explored under different real time environment. Also, it has been observed that various feature extraction techniques are suitable differently for different bit rates and at different noise levels.

## 3. EXPERIMENTAL SETUP

## 3.1 Data Collection

Speech has been recorded by using Audacity toolkit. These speech recordings have been performed in the real time environment. Real time environment consist of the vacant room with no noise, with noise of fan or any other random noise. These recordings have been recorded under particular pattern and are stored in chunks. For instance, in order to understand the need of recordings for experiments, Author understood the impact of noise and other distorted signals which cannot be considered as good speech signal. Also, for recordings, microphones and headphones have been used. The microphone used is of HP (dual microphone and headphone). 500 female speech samples were considered for each word in a particular environment. Testing samples were 100 for each word.

## 3.2 Bit Rates

In this work, different bit resolutions which have been considered are 8 bit, 16 bit and 24 bit. Initially, telephonic (landline) signals used 8 bit resolution at 8000 Hz of sampling rate which computes 64 kbps bit rate. The speech signals are then tested over this bit rate. Next, in order to improve the digitization, 16000 Hz are used instead of 8000 Hz for betterment. However, for radio broadcasting, 16 bit recordings sampled at 16000 Hz are being used. Further this has been improved by introducing 16 bit 44100 Hz recordings which are used in compact discs. As this work is based on analysis of speech signals at different, we use different bit rates for the same. The noise produced by human auditory system theoretically varies up-to 16000 Hz, however it cannot produce more than 7.5 kHz. Also, the audio signals heard by human being are between 20 Hz and 20000 Hz. According to Nyquist theorem, 16 bit recordings sampled at 44100 Hz are sufficient enough for speech recognition. Moreover, more the number of bits, more will be quality and lesser will be compression

## 3.3 Workflow

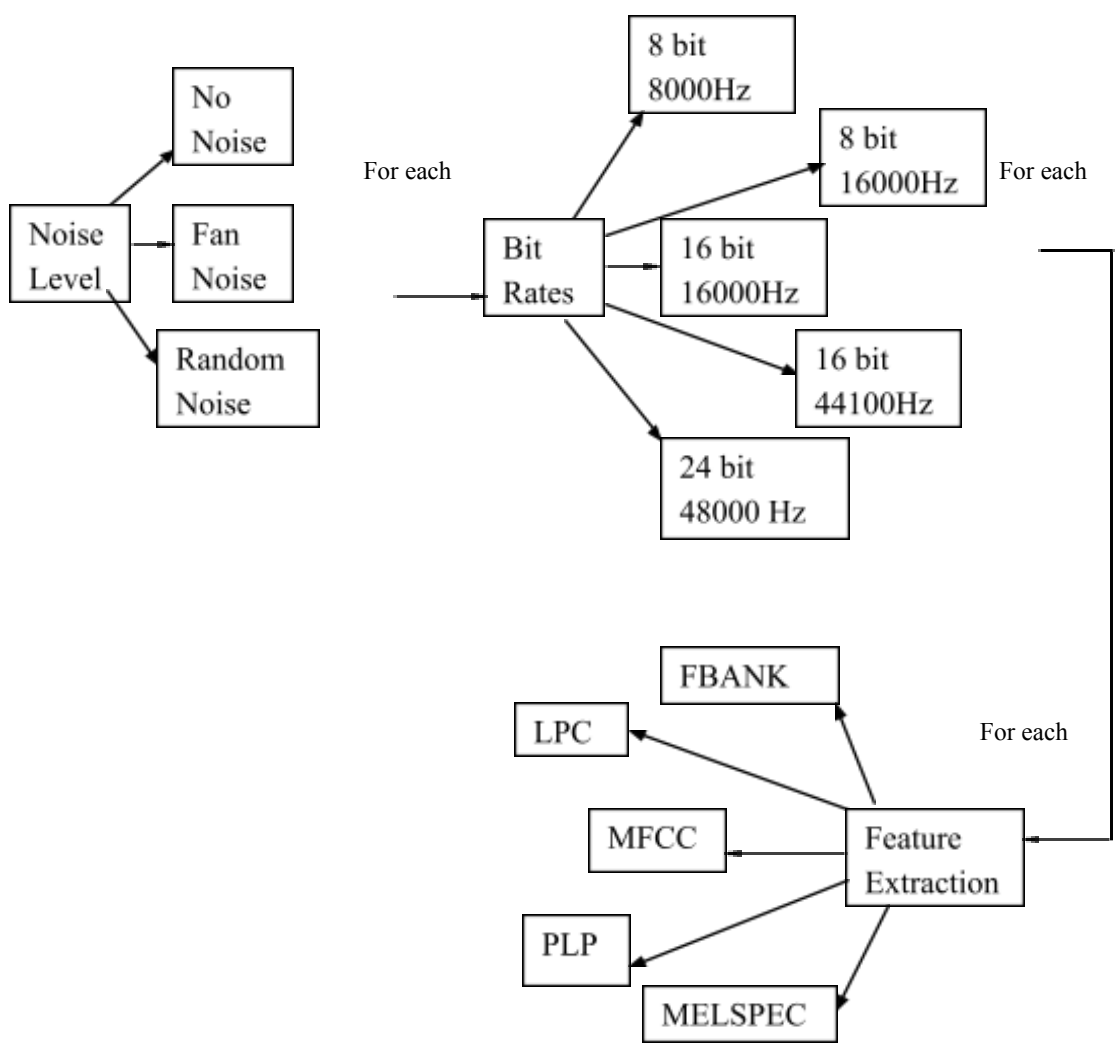

**Fig 1: Flow chart of experimental setup**

Data has been recorded for different set differently. 500 speech samples have been recorded for each dataset. More precisely if methodology is described, it has been observed that speech recording for three noise levels is recorded:

- without noise
- fan noise
- random noise

Further, each one has been classified as different bit rates:

- 8 bit 8000 Hz Mono PCM signal
- 8 bit 16000 Hz Mono PCM signal
- 16 bit 16000 Hz Mono PCM signal
- 16 bit 44100 Hz Mono PCM signal
- 24 bit 48000 Hz Mono PCM signal

Finally, for each type of noise and for each bit rate, five different feature extraction techniques have been extracted:

- Mel Frequency Cepstrum Coefficients
- Linear Predictive Coefficients
- Perceptual Linear Predictions
- Filter banks
- Mel spectrum

# 4. OBSERVATION

Speech Corpus obtained by recording speech samples at 24 bit resolution sampled at 48000 Hz in order to convert the analog signal into digital one, has been used as trained data. This data is trained at mono channel using Audacity and set of testing samples on same bit rate have been used to perform operations. The observation for 24 bit 48000 Hz testing samples on the given speech corpus have been observed and outputs have been prepared in Table 2.

**Table 2: Testing of samples on speech corpus recorded in vacant room**

| Digit | MFCC | LPC | PLP | FBANK | MELSPEC |
|-------|------|-----|-----|-------|---------|
| One | Correct | Correct | In-Correct | Correct | Correct |
| Two | Correct | In-Correct | Correct | Correct | In-Correct |
| Three | In-Correct | Correct | In-Correct | Correct | In-Correct |
| Four | Correct | In-Correct | Correct | In-Correct | Correct |
| Five | Correct | Correct | Correct | In-Correct | Correct |
| Six | Correct | Correct | Correct | Correct | Correct |
| Seven | Correct | Correct | Correct | In-Correct | In-Correct |
| Eight | Correct | Correct | Correct | Correct | Correct |
| Nine | Correct | Correct | In-Correct | Correct | In-Correct |
| Zero | Correct | Correct | Correct | Correct | In-Correct |
| Percentage | 90% | 80% | 70% | 70% | 50% |

It has been observed that MFCC feature vectors perform the best results in absence of noise levels. However, MELSPEC has proved to be least efficient technique for feature extraction and recognition. Thereafter, noise level has been used in order to observe the digits in presence of fan noise. Following observations have been made for the digits mentioned above in presence of fan noise at highest speed.

**Table 3: Testing of samples on speech corpus recorded in vacant room with Fan noise**

| Digit | MFCC | LPC | PLP | FBANK | MELSPEC |
|-------|------|-----|-----|-------|---------|
| One | Correct | Correct | In-Correct | In-Correct | Correct |
| Two | Correct | Correct | Correct | In-Correct | In-Correct |
| Three | In-Correct | Correct | In-Correct | Correct | In-Correct |
| Four | Correct | In-Correct | In-Correct | In-Correct | Correct |
| Five | Correct | In-Correct | Correct | Correct | In-Correct |
| Six | Correct | Correct | Correct | Correct | In-Correct |

| Seven | Correct | Correct | Correct | In-Correct | In-Correct |
|---|---|---|---|---|---|
| Eight | In-Correct | In-Correct | In-Correct | Correct | Correct |
| Nine | Correct | Correct | Correct | In-Correct | Correct |
| Zero | In-Correct | In-Correct | Correct | Correct | Correct |
| Percentage | 70% | 60% | 60% | 50% | 50% |

It has been observed in Table 3, that the speech recognition in presence of fan has been used and subsequently the accuracy of correctness has been reduced. This is certainly due to introduction of noise level. Consequently, another random noise has been used. This noise has been recorded at same features which have been used for original training set.

**Table 4: Testing of samples on speech corpus recorded in room with random noise**

| Digit | MFCC | LPC | PLP | FBANK | MELSPEC |
|---|---|---|---|---|---|
| One | Correct | Correct | Correct | Correct | Correct |
| Two | Correct | Correct | Correct | Correct | In-Correct |
| Three | In-Correct | Correct | Correct | Correct | In-Correct |
| Four | Correct | In-Correct | In-Correct | In-Correct | In-Correct |
| Five | Correct | Correct | Correct | Correct | Correct |
| Six | Correct | Correct | In-Correct | In-Correct | Correct |
| Seven | Correct | Correct | Correct | Correct | In-Correct |
| Eight | Correct | In-Correct | Correct | Correct | In-Correct |
| Nine | Correct | Correct | Correct | Correct | Correct |
| Zero | In-Correct | Correct | In-Correct | In-Correct | Correct |
| Percentage | 80% | 80% | 70% | 70% | 50% |

## 4.1 Result Analysis

Overall Graphical Representation of comparison of various feature extraction techniques in vacant room: without considering any external noise, in the presence of fan noise and in presence of random noise. Initially, no noise has been introduced in speech samples for any of the bit rates. The results have been shown in fig 2.

However, these statistics may change for the voice observed in the presence of fan noise. Experiments and observations helped to determine the same as shown in fig 3.

As the results are analyzed for different bit rates sampled at different frequency levels, it has been observed that MFCC feature vector techniques have proved to be the best feature extraction technique irrespective of bit rates. Hence, MFCC gives more/ equal accuracy as compared to other techniques. However, LPC technique is comparably good enough. Fbank is still better than PLP at times which in turn is better than Melspec.

It gives the impact of almost same observations except that FBank clearly proves better than PLP recognition technique. Also, it has been observed that the recognition at 24 bit 48000 Hz is better than 16 bit 44100 Hz bit rate which in turn is better than other lower bit rated samples. Overall, MFCC has proved to be the best possible technique used for speech recognition. Further, random noise is used to identify the problems faced during recognition of utterances in its presence. The same recordings which have been recognized earlier are embedded with random noise at different bit rates.

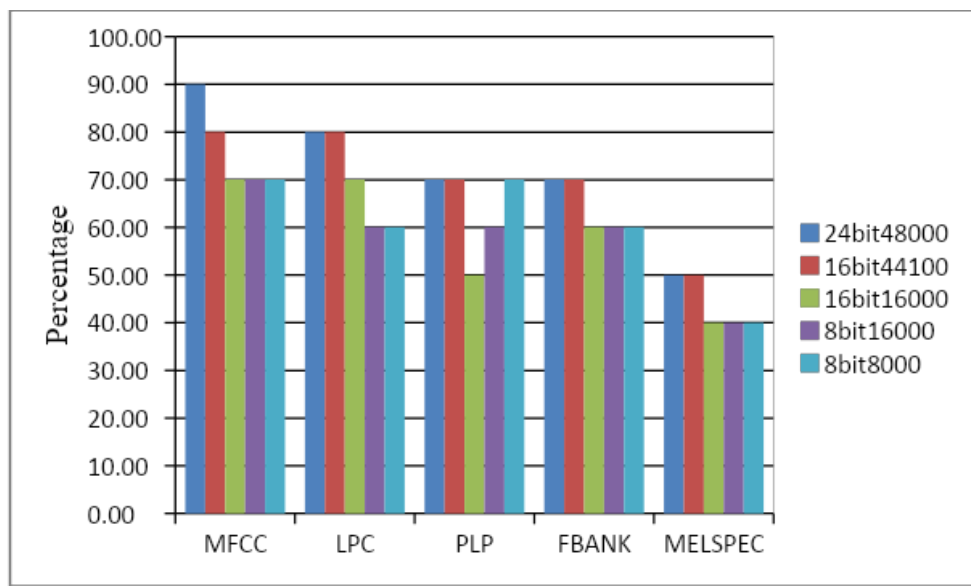

**Fig. 2: Comparison of speech recordings without any noise**

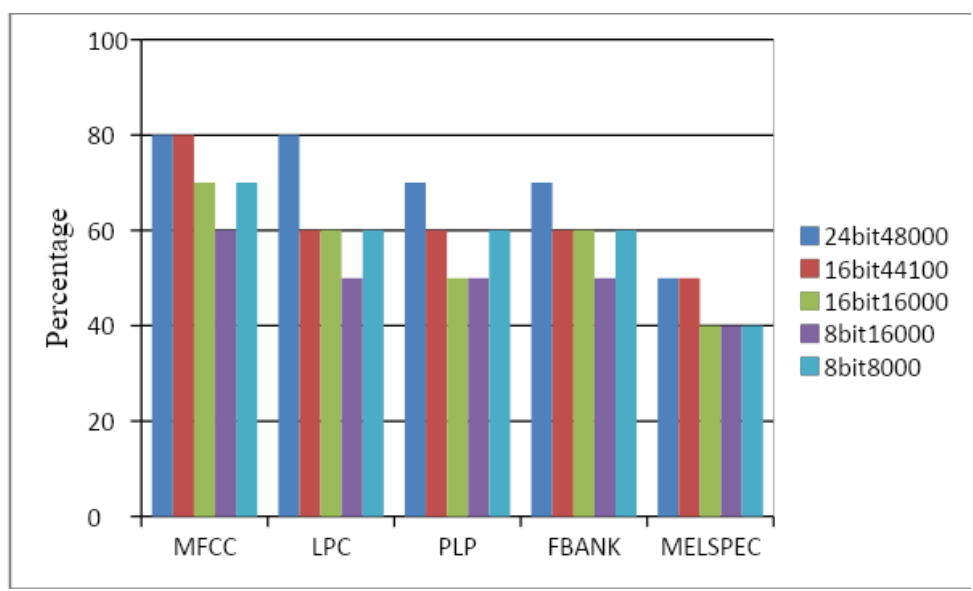

**Fig. 3: Comparison of speech recordings in presence of noise of Fan**

The recognition experiment performed on this dataset is observed as in fig 4. Finally, it has been observed that 24 bit 48000 Hz bit rate has been proved to be the best speech recognition bit rate using MFCC feature extraction techniques. However, for 16 bit 44100 Hz has been set as sufficient bit rate to idealize the work to be better for MFCC feature extraction technique.

Hence, presence of random noise affects more than Fan air does. Finally, for given dataset it has been observed that in terms of betterment of accuracy in decreasing order,

MFCC > LPC > FBANK > PLP > MELSPEC

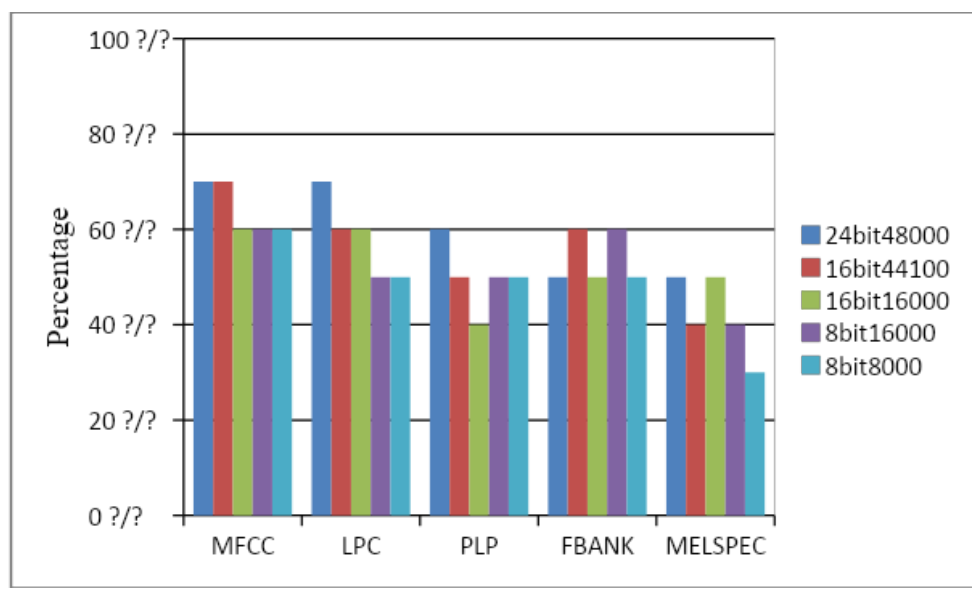

**Fig. 4: Comparison of speech recordings in presence of random noise**

## 5. CONCLUSION

It can be concluded that as we lower the sampling rate and bit rate, the clarification of speech utterances decreases and cannot be recognized clearly. It has been observed that MFCC feature extraction technique is the best suited technique among all the other feature extraction technique for isolated digit recognition. The best environment suited for real time handling of speech is 'no noise' environment. Further, it depends upon the noise level which is being introduced to speech signals. However, this is the major concern in the most upcoming field known as big data which is facing storage problem these days. Recent research and development are related to the fact that recognition at lower sampling rates can be made feasible and improved thereby. Also, larger datasets of different people can be carried out to enhance this work in future.In future, other than this initiative, many other tasks can be performed other than this.

- As research has been carried out for limited dataset, it can be expanded to better level by using larger dataset for Isolated words.
- Work can be carried out on connected words, other isolated words or continuous/ spontaneous speech for future research.
- This work has been carried out for few noise levels; however, noise levels may differ depending upon the patterns and application in which this is used.
- There may be other newly explored or hybrid feature extraction techniques which have not been considered in this study.
- Effects of dialects can be observed
- It can be used for voice commands given by person for ATM operations.
- Instead of words, recognition of different phonemes can be used as bi-phonemes or tri-phonemes etc.
- Other models can be used for the same work namely Neural Networks, Dynamic Time Wrapping etc.
- Other bit rates with combination of other sampling rates can be used for better analysis as per requirement.
- The noise level can be measured in decibels and can be considered differently for different environment to make further analysis. The noise level plays pivotal role in evaluation of results.

Also, it can be observed that MFCC is the best feature extraction technique in the presence of real time environment in vacant room which was recorded using Audacity toolkit at mono-signal for various bit rates using HP microphones for female voices. HTK toolkit is used for feature extraction and HMM model is used for recognition purpose. However, it may vary for different situations and for larger dataset.

## References

[1] R. Furlan, "Build your own Google glass: A wearable computer that displays information and records video", *Spectrum,* IEEE, vol. 50, no. 1, pp. 20–21, January 2013.

[2] L. Rabiner and R. Schafer, "Introduction to digital speech processing", *Foundations and Trends in Signal Processing*, Journal of ACM vol. 1, no. 1-2, pp. 1–194, 2007.

[3] L. Rabiner and B. H. Jaung, *Fundamentals of Speech Recognition,* Englewood Cliffs, NJ: Prentice-Hall, 1993.

[4] L. A. Khan, M. S. Baig and A. M. Youssef, "Speaker Recognition from encrypted VoIP communications", *Proceedings of Digital Investigation*, Journal of Elsevier, vol. 7, no. 1-2, pp. 65-73, October 2010.

[5] X. Huang, J. Baker and R Reddy, "A Historical Perspective of Speech Recognition", *Communications of the ACM*, vol. 57, no. 1, January 2014.

[6] S. Furui, "50 years of progress in speech and speaker recognition", *Proceedings of SPECOM 2005,*Patras, pp. 1-9, 2004.

[7] K. H. Davis, R. Biddulph and S. Balashek, "Automatic recognition of spoken digits," *J.A.S.A.*, vol. 24, no. 6, pp. 637-642, 1952.

[8] S. C. Sajjan and C. Vijaya, "Comparison of DTW and HMM for isolated word recognition", *Proceedings of International Conference on Pattern Recognition, Informatics and Medical Engineering (PRIME)*, IEEE, pp. 466-470, 2012.

[9] B. S. Atal and S. L. Hanauer, "Speech Analysis and Synthesis by Linear Prediction of the Speech Wave", *The Journal of the Acoustical Society of America*, IEEE, vol. 50, no. 2, pp.637-655, 1971

[10] M. J. Coker and S. F. Boll, "An improved isolation word recognition system based upon the linear prediction residual", *Acoustics, Speech, and Signal Processing, IEEE International Conference on ICASSP '76*, vol. 1, pp. 206 – 209, 1976

[11] L. R. Rabinar and M. R. Sambur, "Voiced-Unvoiced-Silence Detection Using the Itakura LPC Distance Measure", *Acoustics, Speech, and Signal Processing, IEEE International Conference on ICASSP '77*, vol. 2, pp. 323-326,

[12] H Sakoi and S Chiba, "Dynamic Programming Algorithm Optimization for Spoken Word Recognition", *IEEE Transactions on acoustics, speech and signal processing*, vol. Assp- 26, no. 1, February 1978.

[13] L R Rabiner, A E Rosenberg and S E Levinson, "Considerations in Dynamic Time Warping Algorithms for Discrete Word Recognition", *IEEE Transactions on acoustics, speech and signal processing*, vol. Assp- 26, no. 6, December 1978.

[14] L R Rabiner, A E Rosenberg, S E Levinson and J G Wilpon, "Speaker-Independent Recognition of Isolated Words Using Clustering Techniques", *IEEE Transactions on acoustics, speech and signal processing*, vol. Assp- 27, no. 4, August 1979.

[15] L. R. Rabinar and M. R. Sambur, "An algorithm for determining the endpoints of isolated utterances", The Bell System Technical Journal, pp. 297-315, 1975.

[16] L R Rabiner, A E Rosenberg, L F Lamel and J G Wilpon , "An Improved Endpoint Detector for Isolated Word Recognition", *IEEE Transactions on acoustics, speech and signal processing*, vol. Assp- 29, no. 4, 1981.

[17] S Furui, "Cepstral Analysis Technique for Automatic Speaker Verification", *IEEE Transactions on acoustics, speech and signal processing*, vol. Assp- 29, no. 2, 1981.

[18] A Buzo, A H Gray, R M Gray and J D Markel, "Speech Coding Based Upon Vector Quantization", *IEEE Transactions on acoustics, speech and signal processing*, vol. Assp- 28, no. 5, 1980.

[19] M A Bush, G E Kopec and N Lauritzen, "Segmentation in Isolated Word Recognition Using Vector Quantization", Acoustics, Speech, and Signal Processing, IEEE International Conference on ICASSP '84, vol. 9, 1984

[20] A H Gray and J D Markel, "Distance measures for speech processing", *IEEE Transactions on acoustics, speech and signal processing*, vol. Assp- 28, no. 5, 1980.

[21] L R Rabiner and J G Wilpon, "A Modified K-Means Clustering A Algorithm for Use in Isolated Work' Recognition", *IEEE Transactions on acoustics, speech and signal processing*, vol. Assp- 33, no. 3, pp. 587-594, 1985.

[22] L R Rabiner and B H Jaung, "An Introduction to Hidden Markov Models", *IEEE ASSP Magazine*, pp. 4-16, January 1986.

[23] L R Rabiner and B H Jaung, "Hidden Markov Models for Speech Recognition", Technometrics, vol 33, no. 3, 1991.

[24] S Axelrod, V Goel, R A Gopinath P A Olsen and K Vishweswariah, "Subspace Constrained Gaussian Mixture Models for Speech Recognition", *IEEE Transactions on speech and audio processing*, vol 13, no. 6, pp.1144-1160, November 2005.

[25] S B Davis and P Mermelstein, "Comparison of Parametric Representations for Monosyllabic Word Recognition in Continuously Spoken Sentences", *IEEE Transactions on acoustics, speech and signal processing*, vol. Assp- 28, no. 4, 1980.

[26] F. Rosdi and R. N. Ainon, "Isolated Malay Speech recognition using Hidden Markov Model", *Proceedings of ICCCE*, May 2008.

[27] J. Psutka, L. Muller and J. V. Psutka, "Comparison of MFCC and PLP Parametrizations in Speaker Independent continuous speech recognition task", Eurospeech 2001, Scandanavia.

[28] A. M. Toh, R. Togneri and S. Nordholm, "Investigating robust features for speech recognition in hostile environment", *Asia Pacific Conference on Communication IEEE*, October 2005.

[29] H. Manabe and Z. Zhang, "Multi-stream HMM for EMG-Based Speech Recognition", *Multimedia Laboratories, NTT Docomo, Kanagawa, Japan.*

[30] Muskan and Naveen Aggarwal, "Punjabi Speech Recognition: A Survey", *Proceedings of ICAET*, May 2014.

[31] P.C. Woodland, C. J. Leggetter, J. J Odell, V. Valtchev and S. J. Young, "The 1994 HTK Large vocabulary speech recognition system", *IEEE*, 1995.

[32] P. C. Woodland, M. J.F. Gales, D. Pye and S. J. Young, "Broadcast News Transcription using HTK", *IEEE*, 1997.

[33] V Tiwari, "MFCC and its applications in speaker recognition", *Proceedings of IJET*, 2010.

[34] Bassam A. Q. Al-Qatab and Raja N. Ainon, "Arabic speech recognition using Hidden Markov Model toolkit (HTK)", *IEEE*, 2010.

[35] M Yanzhou and Y Mianzhu, "Russian Speech Recognition System Design Based on HMM", *Proceedings of LEMCS*, 2014.

[36] K. M Krishna, M V Lakshmi and S. S Lakshmi, "Feature Extraction and Dimensionality Reduction using IPS for Isolated Tamil Words Speech Recognizer", *International Journal of Advanced Research in Computer and Communication Engineering*, Vol. 3, Issue 3, March 2014.

[37] S Young, "The HTK Book", *Cambridge University Engineering Department.*

[38] E Vozarikova, J Juhar and A Cizmar, "Dual Shots Detection", *Information and Communication technologies and services*, Vol. 10, Issue. 4, 2012.

[39] M Alsulaiman, G Muhammad and Zulfiqar Ali, "Comparison of Voice Features for Arabic Speech Recognition", *IEEE*, 2011.

[40] H. G. Hirsch, K. Hellwig and S. Dobler, "Speech Recognition at Multiple Sampling Rates", *Eurospeech*, 2001.

[41] C Sanderson and K K. Paliwal, "Effect of different sampling rates and feature vector sizes on speech recognition performances", *IEEE TENCON*, 1997.

[42] Eric W.M. Yu and Cheung-Fat Chan, "Varying bit rate MBelp speech coding via V/UV distribution dependent spectral quantization", *IEEE*, 1997.

[43] Ravi Kumar K M and Ganesan S, "Comparison of Multidimensional MFCC Feature Vectors for Objective Assessment of Stuttered Disfluencies", *Int. J. Advanced Networking and Applications*, Vol. 2, Issue 5, Page No. 854-860, 2011